\documentclass[a4paper,10pt]{article}
\usepackage[cp1250]{inputenc}
\usepackage[T1]{fontenc}
\usepackage{makeidx}
\usepackage[usenames,dvipsnames]{color}
\usepackage{latexsym}
\usepackage{graphicx}
\usepackage{amsmath}
\usepackage{amssymb}
\usepackage{url}
\usepackage{ctable}
\usepackage{subfigure}
\usepackage{geometry}

\begin{document}
\title{Deciding of HMM parameters based on number of critical points for gesture recognition from motion capture data}
\author{Michał Cholewa \and Przemysław Głomb}
\maketitle

\begin{abstract}

This paper presents a method of choosing number of states of a HMM based on number of critical points of the motion capture data. 
The choice of Hidden Markov Models(HMM) parameters is crucial for recognizer's performance as it is the first step of the training and cannot be corrected automatically within HMM. In this article we define predictor of number of states based on number of critical points of the sequence and test its effectiveness against sample data.

%
\end{abstract}

\section{Introduction}
Hidden Markov Models (HMMs) are presently a popular method for recognition of patterns in data sequences, especially time sequences. Their efficiency, however, depends on a number of parameters that have to be decided a priori, before a HMM is trained for recognition.

One of such parameters is the number of states --- HMM is constructed with pre-defined number of states, which does not change later, during the training process. This number is a factor that has an effect both on the constructed HMM detection ratio and also --- on HMM's complexity. It is important to choose the well performing number of states, both giving sufficient detection rate and not generating HMM of too high complexity.

Most often used method is to try out several number of states and decide the one best performing --- this is, however, a time consuming task, influencing the time cost of HMM construction. 
Because of that, a method to estimate those parameters prior to constructing HMM, so it gives good results would be a desirable addition to the process. 

In this article we consider data from motion capture sensors. For these data, we construct predictors of HMM states number based on median value of critical points number in training sequences. We then define measure of effectiveness of such predictor based on Akaike's Information Criterion and then test the predictors against defined measure.

This paper is organized as follows: in section \ref{sec:relatedwork} we discuss related work concerning both HMM use and critical points use in recognition process. Section \ref{sec:HMMS} contains brief introduction to Hidden Markov Models. Section \ref{sec:Approach} describes our approach --- both to constructing predictor basis and how to test its quality, and Section \ref{sec:Experiments} will present our experiments and results.

\section{Related work}\label{sec:relatedwork}

Hidden Markov Models (HMMs) as a method of modelling of data sequences have been well developed and frequently used. While they were at first most often used in the area of cryptography, their application fast widened. They are nowadays a method of choice for speech recognition systems (see e.g. \cite{18}) and used in such areas as protein classification and alignment (as in \cite{19}) and gesture classification (see \cite{4}). 

Use of multi-dimensional HMMs in gesture recognition is described in \cite{1}, where the classification is based upon the input form digital camera. Wilson and Bobick in \cite{2} describe use of parametric HMMs as well as online learning for gathering more gesture executions. 

Gesture recognition has been, mainly, approached as a problem of classification of distinct gestures captured by digital camera as a sequence of images. There are relatively small number of works assuming different way of acquiring data, such as motion capture input. In \cite{22}, a database is presented, that contains input from motion capture gloves for significant number of executions of 22 gestures. The gestures from provided database have been analysed for separability in \cite{21} and application of HMMs on such input data has been successfully tested in \cite{4}. In \cite{5} a critical points approach is used for classification and proven effective.

Given predefined priors (such as number of states) there are a number of ways to build HMM. Many new ideas have been introduced since traditional Baum-Welch algorithm (described in \cite{20}). Some  begin construction with a number of states (either high or low) and then either increase number of states or decrease it towards predefined prior. Specific algorithms for state-splitting \cite{8} and state-merging (such as Bayesian model merging, \cite{12}) have been designed (i.e. Gaussian splitting-merging algorithm tested in speech recognition systems in \cite{11}). In \cite{3} Viterbi Path Counting algorithm was tested, and has been proven effective next to traditional Baum-Welch algorithm of HMM training.

Number of HMM states is an element of topology to be decided a priori, and then remain unchanged during the learning phase. Since priors like the number of states in HMMs are factors deciding of its effectiveness, it has been so far approached in several ways. Most popular ways are greedy algorithms, trying out several possibilities and choosing the one with best results. However this method gives results it tends to be very time consuming. There have been some ways of dealing with the problem. Bakis modelling suggests choosing the number of states corresponding to the length of input sequence (so every datapoint has corresponding state in HMM), what tends to produce HMMs with very large number of sequences. In \cite{14} authors propose choosing between constant number and --- alternatively --- a number depending of length of feature vector and achieved visible effect, in \cite{18} Bakis modelling is improved by its iterative application and more sophisticated dependency function. This method, however more effective in state number estimation than Bakis modelling, requires a large number of computations. In \cite{1} Bakis model is used for gesture recognition to some effect.

\section{Hidden Markov Models} \label{sec:HMMS}

Our input data (e.g. a recorded gesture) is a sequence of symbols $O=O_1O_2\ldots O_T$, where each $O_i$ is a member of the alphabet set $O_i \in \mathcal{L} = \{L_1, \ldots, L_k\}$. For motion capture data, $O_i$ could correspond to result of Vector Quantization of sensor measurements.

We model a set of input data (sequences from alphabet $\mathcal{L}$, e.g. several recordings of a defined gesture)  with a discrete Hidden Markov Model, defined as
\begin{equation}
\lambda = (\mathcal{S}, T, b, \mathcal{L} , E),
\end{equation}
where $\mathcal{S} = \{S_1, \ldots, S_n\}$ is a set of states, $T \in \mathbb{R}^{n \times n}$ a stochastic\footnote{Where as \emph{stochastic} we understand a matrix $Z_{ij} \in \mathbb{R}^{t \times l}$ where $\sum_{i = 1}^l Z_{in} = 1, \forall n = 0, \dots, t$ and $M_{ij} \geq 0$.} transition matrix, $b$ probability vector representing the probability distribution of starting state and $E \in \mathbb{R}^{n \times k}$ a stochastic emission matrix.

Given a HMM $\lambda$ and a sequence $O$ we can compute the log likelihood $\log(p(O)|\lambda)$; a logarithm of probability that $O$ was generated by $\lambda$. This can be done e.g. with Forward Algorithm \cite{18}. Given a set of HMMs $\{\lambda_i\}$ we can use the likelihood to determine the most probable HMM associated with the sequence from
\begin{equation}
\mathop{\mathrm{argmax}}_{i}\log(p(O)|\lambda_i).
\end{equation}
If each $\lambda_i$ corresponds to a different gesture, we can use this to recognize the sequence as one of those gestures.

A standard way to build a HMM model given a number of states $n$ and a set of reference sequences  $\mathcal{T} = \{ O_1, \ldots, O_m \}$ is to use a Baum-Welch algorithm \cite{18}, that iteratively maximizes the likelihoods of sequences from $\mathcal{T}$. The number of states $n$ for this approach must be known beforehand.

Performance of HMM depends both on the set of reference sequences $\mathcal{T}$ and number of its states $n$. While set of references is usually provided beforehand, number of states must be estimated. The choice of $n$ determines both how effectively HMM will recognize sequences and what will be its model complexity (number of parameters). Estimating $n$ is therefore a matter of balancing between model complexity and fit to the data 

The standard approach in this case is to use information criteria (such as AIC --- Akaike Information Criterion) This approach, however, de facto is a greedy algorithm that needs to build several HMMs with different $n$ and find out which of them is the best in context of selected information criterion. It creates a need for an algorithm to select $n$ without time consuming testing.



\section{Our Approach} \label{sec:Approach}
It is our goal to show an effective method of choosing the number of detecting HMM states for input data in form of matrices $M \in \mathbb{R}^{m \times n}$. Our thesis is that number of HMM states corresponding to number of critical points of dataset in question is a viable method of estimating HMM parameters. As \emph{critical points} we will understand points at the end of the sampling interval, local maxima and local minima of data sequence.

\subsection{Input}

Let $\mathcal{I} = \{1, \ldots, I \}$ be a set of class indices, $\mathcal{K} = \{ 1, \ldots, K \}$ be the set of class example indices and let $\{G^{i,k}\}_{i \in \mathcal{I}, k \in \mathcal{K}}$ be a set of matrices where $G^{i,k} \in \mathbb{R}^{J \times l(i,k)}$. We will understand $G^{i,k}$ as the $k$-th example of matrix from class $i$. In context of motion capture data it represents single execution of a gesture of class $i$. Since we consider motion capture data, we have to assume that each matrix $G^{i,k}$ has different number of columns, which represent sequences length. We denote this number as $l(i,k)$.  For each matrix $G^{i,k}$ we will treat it as an ordered set of sequences (matrix' rows) $G_j^{i,k}, j \in \mathcal{J}$ of length $l(i,k)$. Therefore each sequence $G_j^{i,k}$ we will be analysing, will be $j$-th row of matrix $G^{i,k}$ where $\mathcal{J} = \{ 1, \ldots, J \}$. 

\subsection{Preprocessing}
With each sequence $G^{i,k}_j$ we proceed as follows. First, the sequence $G_j^{i,k} $is polynomially interpolated and resampled to length $M$. Next step is normalization of $G_{j}^{i,k} = \left( G_{j,m}^{i,k} \right)_{m = 1}^{M}$ ($m$-th element of $j$-th row of $G^{i,k}$). For a sequence $G_{j}^{i,k}$, let
\begin{equation}
\mu = \frac{\Sigma_{m = 1}^{M} G_{j,m}^{i,k}}{M}
\quad
\sigma = \sqrt{\frac{\sum_{m = 1}^{M} (G_{j,m}^{i,k} - \mu)^2}{M}},
\end{equation}
 $\mu$ being a mean value of given sequence, and $\sigma$ --- its standard deviation. The normalization of sequence $\left( G_{j,m}^{i,k}\right)_{m=1}^{M}$ can be represented as the transformation $P: \mathbb{R} \rightarrow \mathbb{R}$
\begin{equation}
P(G_{j,m}^{i,k})=
\frac{G_{j,m}^{i,k} - \mu}{\sigma}.
\end{equation}
In other words, each value $G_{j,m}^{i,k}$ of input sequence is mean shifted and standard deviation normalized. From this point onward we will understand $G_{j}^{i,k}$ as sequences of normalized values.

\subsection{Computing the number of critical points}
Next, we calculate the critical points --- by which we understand end points of the sequence and local extrema. 
We consider a point $G_{j,m}^{i,k}$ to be a local maximum, if its value is highest in the interval $\langle m - \gamma, m + \gamma\rangle$ where $\gamma \geq 1$. If $m - \gamma < 1$ or $m + \gamma > M$, we pad first or last value of the sequence appropriately.  Local minima are calculated in similar way.
For a sequence $G_{j}^{i,k}$, we denote number of local maxima as $\text{cp}^{\text{max}}(G_{j}^{i,k})$ and number of local minima as $\text{cp}^{\text{min}}(G_{j}^{i,k})$. Considering the beginning and end of a sequence, total number of critical points equals
\begin{equation}
\text{cp}(G_{j}^{i,k}) = \text{cp}^{\text{max}}(G_{j}^{i,k}) + \text{cp}^{\text{min}}(G_{j}^{i,k}) + 2.
\end{equation}

Computed value $\text{cp}(G_{j}^{i,k})$ is the base to our predictor. It is our thesis, that predictor close to $\text{cp}(G_{j}^{i,k})$ gives good results when deciding the number of states of HMM to recognize $G_{j}^{i,k}$.

\subsection{Test}
To test our thesis, we need to observe, how predictor acts when used in practice. We decided to measure quality of our predictor in context of Akaike Information Criterion, which merges both effectiveness and computational complexity. To do this we propose the following test.
\subsubsection{Clustering of values}
Having constructed a set of normalized sequences of equal length, we now proceed to clusterization. First, k-means clustering method is used to partition values $G_{j,m}^{i,k}, i\in I, j \in \mathcal{J}, k \in \mathcal{K}, m=1, \ldots, M$ into $c \in \mathcal{C}$ clusters, where $\mathcal{C} = \{ c_l,c_l +1, \ldots, c_h \}$. 

Therefore, after this step, each sequence $G_{j}^{i,k}$ generates $c_h -c_l + 1$ sequences $G_{j}^{i,k,c}$ consisting of positive integers (index of cluster).

\subsubsection{HMM construction and computation of AIC coefficient for each sequence}

For HMM construction, we take all $G_{j}^{i, k ,c}$ and group them by $k$ --- in other words for each sensor of each gesture and given number of clusters we take all executions and put them in one set. Now, every gesture $i$ and every sensor $j$, for chosen number of clusters $c$ we group all sequences $\{G^{i,k,c}_{j}\}_{k \in \mathcal{K}}$. To each of these sets we assign $\text{cp}(G_j^{i})$, being the median of set $\{\text{cp}(G_{j}^{i,k})\}_{k \in \mathcal{K}}$, which is independent of number of clusters $c$. Then we create pairs $F_{ijc} = \left(\{G^{i,k,c}_{j}\}_{k \in \mathcal{K}}, \text{cp}(G_j^{i})\right)$.

Pairs $F_{ijc}, i \in \mathcal{I}, j \in \mathcal{J}, c \in \mathcal{C}$ will be the data used to HMM construction. For each pair $F_{ijc}$ we construct $({st}_h-{st}_l + 1)$ HMMs referred to as $\lambda(F_{ijc}, n)$ where $n= {st}_l,\dots, {st}_h $ is the number of states.

Then, using standard method for each $\lambda(F_{ijc}, n)$, we compute logarithm of probability for each sequence $G_{j}^{i,k,c}$ from $F_{ijc}$. That means, for each $G_{j}^{i,k,c}$ we acquire $({st}_h-{st}_l + 1)$ logarithms of probability $\log p(G_{j}^{i, k ,c}, \lambda(F_{ijc}, n))$, where $ n= {st}_l,\dots, {st}_h $.  

To rate those results we consider two factors: logarithm of probability $\log p(G_{j}^{i,k,c}, \lambda(F_{ijc}, n))$ and complexity cost $q$. The computational complexity of generating result with HMM of $n$ states is $(n^2 + na)$, since such a HMM operates on transition matrix $T \in \mathbb{R}^{n \times n}$ and emission matrix $E\in \mathbb{R}^{n \times k}$.

There are many criteria that include both of these factors, and in our approach we use Akaike Information Criterion.
We compute the Akaike information criteria value for a set $F_{ijc}$ and number of states $n$, as follows:
\begin{equation}
\mathrm{AIC}(F_{ijc}, n) = -2\sum_{k=1}^{K} \log p(G_{j}^{i,k,c}, \lambda(F_{ijc}, n))+2q,
\end{equation}
where $q = n^2$ (representing the size of the transition matrix used by HMM). Separate $\mathrm{AIC}(F_{ijc}, n)$ value is computed for each number of HMM states $n$.

\subsubsection{Positioning}

We want to observe if the number of critical points $\text{cp}(G_j^i)$ can be used to build a good predictor of the number of states $n$ of the model with the lowest $\mathrm{AIC}$ value (preferable model). To do that, for each $F_{ijc}$ and for a given predictor $cp$ we will calculate a value $\xi(ijc,cp) \in \langle 0,1 \rangle$ which will aid us to rate different predictors.

For given range of $n= {st}_l , \ldots, {st}_h$, $F_{ijc}$ and predictor $cp$  we define three values:
\begin{itemize}
\item $\mathrm{AIC}(F_{ijc})_{\min} = \min_{n={st}_l,\ldots,{st}_h}\mathrm{AIC}(F_{ijc}, n)$ --- the lowest value of $\mathrm{AIC}$ given set $ F_{ijc} $;
\item $\mathrm{AIC}(F_{ijc})_{\max} = \max_{n={st}_l,\ldots,{st}_h}\mathrm{AIC}(F_{ijc}, n)$ --- the highest value of $\mathrm{AIC}$ given set $ F_{ijc} $;
\item $\mathrm{AIC}(F_{ijc})_{cp} = \mathrm{AIC}(F_{ijc}, cp)$ --- the value of $\mathrm{AIC}$ for number of HMM states equal to a our predictor $cp$.
\end{itemize}

We define a measure of similarity of $\mathrm{AIC}(F_{ijc})_{cp}$ to the minimum as
\begin{equation}
\xi(ijc, cp) = \frac{\mathrm{AIC}(F_{ijc})_{cp}-\mathrm{AIC}(F_{ijc})_{\min}}{\mathrm{AIC}(F_{ijc})_{\max}-\mathrm{AIC}(F_{ijc})_{\min}}.
\end{equation}

It is easy to see, that $\xi(ijc, cp) \in \left<0,1 \right>$, for given $\mathrm{AIC}(F_{ijc})_{\max}$,  $\xi(ijc, cp) \rightarrow 0$ as $\mathrm{AIC}(F_{ijc})_{cp} \rightarrow \mathrm{AIC}(F_{ijc})_{\min}$. Also $\xi(ijc, cp) \rightarrow 1$ when $\mathrm{AIC}(F_{ijc})_{cp} \rightarrow \mathrm{AIC}(F_{ijc})_{max}$. 

Therefore, we conclude that if $\xi(ijc, cp)$ is close to zero, the $\mathrm{AIC}(F_{ijc})_{cp}$ is near $\mathrm{AIC}(F_{ijc})_{\min}$, which means $cp$ gives us close estimate of number of HMM states, that produces minimal value of Akaike Information Criterion. Therefore $cp$ is a good predictor of $n$ in terms of Akaike Information Criterion.

\section{Experiments}\label{sec:Experiments}

The objective of the experiments is to determine if predictor $\text{cp}$ based on number of critical points in detected sequences $G_j^{i,k}$ is a good predictor for HMM number of states in terms of Akaike information Criterion. To do that we will calculate values of $\xi(ijc, \text{cp})$ for analyzed $F_{ijc}$ and aggregate average value of 
\begin{equation}
\xi(\text{cp}) = \sum_{i \in \mathcal{I}, j \in \mathcal{J}, c \in \mathcal{C}} \frac{\xi(ijc, \text{cp})}{IJ({cl}_h - {cl}_l + 1)}.
\end{equation}

It is easy to see, that the lower $\xi(\text{cp}) \in  \langle 0,1 \rangle$ is, the better predictor $cp$ is.

For experiments we used motion capture glove, which produces finite sequence of 11-dimensional real vectors, representing the readings of 10 installed sensors (5 finger bend sensors, 2 accelerometers ``pitch'' and ``roll'' and 3 accelerometers ``OX'', ``OY'' and ``OZ'' recording movement of the hand) and time coordinate. While average number of critical points remains similar in all gestures (see Table \ref{table:avgcpges}), it varies when grouping $F_{ijc}$ by sensor ($j$) --- especially  accelerometers ``OX'',``OY'' and ``OZ'' have very high number of critical points, while finger sensors show significantly smaller averages, as we can observe in Table \ref{table:avgcpsen}

\ctable[
  caption = Average number of critical points by gesture, 
  label = table:avgcpges,
  pos = h
]{crcr}
{}{\FL 
 Gesture & Avg. number of critical points & Gesture & Avg. number of critical points \ML
 1 & 6.4 & 11 & 6.6 \NN
 2 & 9.2 & 12 & 12.9 \NN
 3 & 8.8 & 13 & 9.5\NN
 4 & 5.9 & 14 & 10.3 \NN
 5 & 6.2 & 15 & 8.5 \NN
 6 & 6.6 & 16 & 6.3 \NN
 7 & 8.9 & 17 & 7.3 \NN
 8 & 8.4 & 18 & 10.0 \NN
 9 & 6.8 & 19 & 10.1 \NN
 10 & 9.4 & 20 & 6.5 \LL
}

\ctable[
  caption = {Average number of critical points by sensor. As we see, accelerometers readings (6-10) have significantly more critical points than the finger sensors}, 
  label = table:avgcpsen,
  pos = h
]{crcr}
{}{\FL 
 Sensor & Avg. number of critical points & Sensor& Avg. number of critical points \ML
 1 & 3.90 & 6 & 7.25 \NN
 2 & 4.80 & 7 & 9.30 \NN
 3 & 4.35 & 8 & 15.90 \NN
 4 & 4.00 & 9 & 14.50 \NN
 5 & 3.55 & 10 & 14.65 \LL
}

As the input data we  used data collected from motion capture glove, of $K = 15$ of $I= 20$ gestures. Each execution $G_i^k$ gives us set of $J = 10$ sequences, one for each of sensors in the glove. These data are resampled to the length of $M=64$. Having discretized sequences of length $64$ we decided for this experiment, that local extrema will be calculated with $\gamma = 1$. It means, a value is consider local maximum or minimum if it is higher or lower ---respectively --- than its neighbours.

We then conducted two experiments:

\begin{itemize}
\item In Experiment A we have set ${cl}_l = 4$ and ${cl}_h = 11$, then used generated $\{F_{ijc}\}_{i \in \mathcal{I}, j \in \mathcal{J} , c \in \mathcal{C}}$ where $\mathcal{I} = \{1, \ldots, 20 \}, \mathcal{J} = \{ 1, \ldots, 10 \}, \mathcal{C} = \{  4, \ldots, 11\}$ as single data set.
\item In Experiment B, we have generated 8 different datasets $\{F_{ijc}\}_{i \in  \mathcal{I}, j \in  \mathcal{J}}$ each one with fixed $c = 4, \ldots, 11$. This experiment is designed to analyse if the efficiency of our predictor varies depending of number of clusters.
\end{itemize}

Both in Experiment A and B we have checked three different predictors $\text{cp}$ for selecting number of states for HMM to detect $F_{ijc}$. We considered
\begin{itemize}
\item all points computed from the sensor data $\text{cp}=\text{cp}(G_j^i)$;
\item all points without the boundary points $\text{cp}=\text{cp}(G_j^i) - 2$;
\item $\text{cp}=\text{cp}(G_j^i) - 1$, which corresponds to the number of trends.
\end{itemize}

Also, in both A and B, given different nature of sensor for fingers ($1, \ldots, 5$) and accelerometers ($6, \ldots, 10$), we have decided that we will consider also two possible ranges of $j$

\begin{itemize}
\item $j = 1, \ldots, 10$ representing all sensors input.
\item $j = 1, \ldots, 5$ representing only fingers sensors.
\end{itemize}

Therefore, both Experiment A and B will be performed with three different predictors and two different sensor ($j$) range.

The result of each experiment is a value $\xi(cp) \in [0,1]$, representing how averagely close to the $\mathrm{AIC}(F_{ijc})_{min}$ was  $\mathrm{AIC}(F_{ijc})_{cp}$. It is our thesis, that HMM with appropriately set number of states will give very low (close to $0$) results.

\subsection{Results}

\subsubsection{Experiment A}

First we will see the results of analysis of all the sequences. In the Table \ref{table:results} we can see the average positioning ratio $\xi$ of $AIC(F_{ijc}, cp)$. We have considered three different predictors and two different ranges of $j$ --- all sensors and only finger sensors.

\ctable[
  caption = Results of analysis of all files of $\xi$ for three different predictors cp. We can observe significantly better results with finger sensors than with accelerometers, 
  label = table:results,
  pos = h
]{l@{\hspace{0.25cm}}c@{\hspace{0.25cm}}c@{\hspace{0.25cm}}c}
{}{\FL
 & $\text{cp}=\text{cp}(G_j^i)$ & $\text{cp}=\text{cp}(G_j^i)-2$ & $\text{cp}=\text{cp}(G_j^i)-1$ \NN
 & (all points) & (no boundaries) & (trends only) \ML
All sensors & 0.2003 & 0.1457 & 0.1723 \NN
Fingers only & 0.0216 & 0.0125 & 0.0174 \LL
}

Additionally, we have computed average $\xi$ of $AIC(F_{ijc}, cp)$ for all gestures and all sensor separately, the results of which we can see in Tables \ref{table:resultsByGesture} and \ref{table:resultsBySensor}.

\ctable[
  caption = Average value of $\xi$ by gesture, 
  label = table:resultsByGesture,
  pos = h
]{cccc}
{}{\FL 
 Gesture & Average $\xi$ & Gesture & Average $\xi$ \ML
 1 & 0.0809 & 11 & 0.0837 \NN
 2 & 0.1913 & 12 & 0.3580 \NN
 3 & 0.1606 & 13 & 0.1681 \NN
 4 & 0.0563 & 14 & 0.2068 \NN
 5 & 0.0624 & 15 & 0.1496 \NN
 6 & 0.0860 & 16 & 0.0640 \NN
 7 & 0.1746 & 17 & 0.1265 \NN
 8 & 0.1597 & 18 & 0.2134 \NN
 9 & 0.0994 & 19 & 0.2016 \NN
 10 & 0.1865 & 20 & 0.0852 \LL
}

\ctable[
  caption = Average value of $\xi$ by sensor. We can observe the quality of the predictor deteriorating when applied towards accelerometers data, 
  label = table:resultsBySensor,
  pos = h
]{cccc}
{}{\FL 
 Sensor & Average $\xi$ & Sensor& Average $\xi$ \ML
 1 & 0.0089 & 6 & 0.0496\NN
 2 & 0.0170 & 7 & 0.1118 \NN
 3 & 0.0164 & 8 & 0.4675 \NN
 4 & 0.0108 & 9 & 0.3818 \NN
 5 & 0.0094 & 10 & 0.3840 \LL
}

\subsubsection{Experiment B}

Then we have the results in groups divided by number of clusters. What was expected, is that larger number of clusters will improve the results (to a certain point). That would indicate that perhaps proposed method is merely an exchange --- instead of problem of choice the number of HMM states we will now face the problem of choosing number of clusters for clusterization. As we see in table  \ref{table:results2}, this did not happen. The value of Akaike Information Criterion does not change significantly when we use varied number of clusters $c$, which suggests that value of the predictor is independent of parameter $c$.

\ctable[
  caption = {Average value of $\xi$ grouped according to number of clusters. As we can see, the values do not change significantly along with change of parameter $c$ which suggests that the quality of predictor cp is independent on number of clusters},  
  label = table:results2
]{l@{\hspace{0.25cm}}c@{\hspace{0.25cm}}c@{\hspace{0.25cm}}c}
{}{\FL
 &  $\text{cp}=\text{cp}(G_j^i)$ & $\text{cp}=\text{cp}(G_j^i)-2$ & $\text{cp}=\text{cp}(G_j^i)-1$ \NN
 & (all points) & (no boundaries) & (trends only) \ML
All sensors, $c = 4$ & 0.178 & 0.134 & 0.156 \NN
Fingers only, $c = 4$  & 0.015 & 0.026 & 0.024 \NN
All sensors, $c = 5$ & 0.175 & 0.132 & 0.155 \NN
Fingers only, $c = 5$  & 0.014 & 0.026 & 0.025 \NN
All sensors, $c = 6$ & 0.174 & 0.133 & 0.155 \NN
Fingers only, $c = 6$  & 0.015 & 0.029 & 0.027 \NN
All sensors, $c = 7$ & 0.174 & 0.133 & 0.155 \NN
Fingers only, $c = 7$  & 0.015 & 0.031 & 0.028 \NN
All sensors, $c = 8$ & 0.173 & 0.133 & 0.154 \NN
Fingers only, $c = 8$  & 0.015 & 0.030 & 0.028 \NN
All sensors, $c = 9$ & 0.167 & 0.129 & 0.149 \NN
Fingers only, $c = 9$  & 0.013 & 0.030 & 0.027 \NN
All sensors, $c = 10$ & 0.173 & 0.133 & 0.154 \NN
Fingers only, $c = 10$  & 0.015 & 0.032 & 0.029 \NN
All sensors, $c = 11$ & 0.174 & 0.133 & 0.155 \NN
Fingers only, $c = 11$  & 0.016 & 0.032 & 0.029 \LL
}

While the results of HMMs being taught with input sequences of course differ, yet the values of AIC coefficients are similar, and expected improvement of results along with rising number of clusters did not occur.

\section{Conclusion}

Choosing number of HMM states in dependence of number of critical points in given dataset gives very good results when dealing with finger sensors, where computed AIC effectiveness of HMMs with number of states equal to $cp$ is averagely in upper 2\% of the results. 
Such high results suggests that efficiency-wise proposed method.
When we apply this method to all sensors, the efficiency of this method drops. Even though it still produces above average results, it is visible, that accelerometers and their readings are considerable challenge.
What was surprising, changing the number of clusters in the clusterization phase did not have significant effect on the method efficiency, which suggests that proposed solution is not simple exchange of problem of selecting number of HMM states to a problem of deciding number of clusters for clusterization.

%
%
%
%
%

\begin{thebibliography}{}
\bibitem{18} L.R. Rabiner, A Tutorial on Hidden Markov Models and selected applications in speech recognition, Proc. IEEE 77(1989) 257-286
\bibitem{19} A. Krogh, B. Brown, I.S. Mjan, K. Sjolander, D. Haussler, Hidden Markov Models in Computational Biology Applications to Protein Modeling, Journal of Molecular Biology 235(1994) 1501-1531
\bibitem{4}  M. Romaszewski, P. G{\l}omb, , The Effect of Multiple Training Sequences on HMM Classification of Motion Capture Gesture Data, Computer Recognition Systems 4 2011
\bibitem{1} J. Yang, Y. Xu, Hidden Markov Model for gesture recognition, Gesture 5(1994) 1-52
\bibitem{2} A.D. Wilson, A.F. Bobick, Hidden Markov Models for Modeling and Recognizing Gesture Under Variation, International Journal of Pattern Recognition and Artificial Intelligence 15(2000) 123-160
\bibitem{22} M. Romaszewski, P. G{\l}omb, S. Opozda, A. Sochan, Choosing and modelling gesture database for natural user interface, Computer Recognition Systems 4 2011
\bibitem{21} M. Romaszewski, P. Gawron, P. G{\l}omb, Towards a natural gesture interface: LDA-based gesture separability, arXiv:1109.5034
\bibitem{5} M. Cholewa, P. G{\l}omb, Gesture Data Modeling and Classification Based on Critical Points Approximation, Computer Recognition Systems 4, 2011
\bibitem{20} L.R. Welch, Hidden Markov Models and the Baum-Welch algorithm, IEEE Information Theory Society Newsletter 53(2003) 10-14
\bibitem{8} M. Ostendorf, H. Singer, HMM topology design using maximum likelihood successive state splitting, Computer Speech Languages 1997 17-41
\bibitem{12} A. Stolcke, A.M. Omohundro, Hidden Markov Model Induction by Bayesian Model Merging, Advances in Neural Information Processing Systems 1993: 11-18  
\bibitem{11} A. Sankar, Experiments with a Gaussian Merging-Splitting Algorithm for HMM Training for Speech Recognition, In Proceedings of the Broadcast News Transcription and Understanding Workshop 1998 99-104
\bibitem{3} N. Liu, B.C. Lovell, Gesture Classification Using Hidden Markov Models and Viterbi Path Counting, VIIth Digital Image Computing: Techniques and Applications 2003
\bibitem{14} S. Gunter, H. Bunke, Optimizing the Number of States, Training Iterations and Gaussians in an HMM-based Handwritten Word Recognizer, Proceedings. Seventh International Conference on Documents Analysis and Recognition 1(2003) 472-476
\end{thebibliography}

\end{document}